\documentclass[conference]{IEEEtran}
\IEEEoverridecommandlockouts
\usepackage{cite}
\usepackage{amsmath,amssymb,amsfonts}
\usepackage{graphicx}
\usepackage{textcomp}
\usepackage{xcolor}
\usepackage{tikz}
\usepackage{multicol}
\usepackage{multirow}
\usepackage[bookmarks=true,hidelinks]{hyperref}
\usepackage[caption=false]{subfig}

\usepackage{algorithm}
\usepackage{algpseudocode}

\def\BibTeX{{\rm B\kern-.05em{\sc i\kern-.025em b}\kern-.08em
    T\kern-.1667em\lower.7ex\hbox{E}\kern-.125emX}}

\graphicspath{ {figures/} }  

\usepackage[caption=false]{subfig}

\usepackage{acronym}
    
\usepackage{cancel}

\usepackage[export]{adjustbox}

\usepackage{soul}

\begin{document}

\acrodef{fl}[FL]{Federated Learning}
\acrodef{sgld}[SGLD]{Stochastic Gradient Langevin Dynamics}
\acrodef{dsgld}[DSGLD]{Decentralized Stochastic Gradient Langevin Dynamics}
\acrodef{ml}[ML]{Machine Learning}
\acrodef{iiot}[IIoT]{Industrial Internet of Things}
\acrodef{nn}[NN]{Neural Network}
\acrodef{vi}[VI]{Variational Inference}
\acrodef{mcmc}[MCMC]{Markov Chain Monte Carlo}
\acrodef{d2d}[D2D]{Device-to-Device}
\acrodef{ece}[ECE]{Expected Calibration Error}
\acrodef{cd-bfl}[CD-BFL]{Compressed Decentralized Bayesian Federated Learning}
\acrodef{cf-fl}[CF-FL]{Compressed Frequentist Federated Learning}

\title{Compressed Bayesian Federated Learning for Reliable Passive Radio Sensing in Industrial IoT
\thanks{This paper is partially funded by the EU in the call HORIZON-HLTH-2022-STAYHLTH-01-two-stage under grant agreement No 101080564 and by the project HORIZON-EIC-2022-PATHFINDEROPEN-01 under grant Agreement No
101099491.}
}

%\author{\IEEEauthorblockN{Anonymous authors}}

\author{\IEEEauthorblockN{Luca Barbieri\textit{$^{1,2}$}, Stefano Savazzi\textit{$^{2}$}, Monica Nicoli\textit{$^{1}$}}
\IEEEauthorblockA{\textit{$^{1}$ Politecnico di Milano, Milan,
Italy}, \textit{$^{2}$ Consiglio Nazionale delle Ricerche, Milan, Italy}
}}
\maketitle

\begin{abstract}
Bayesian Federated Learning (FL) has been recently introduced to provide well-calibrated Machine Learning (ML) models quantifying the uncertainty of their predictions. 
Despite their advantages compared to frequentist FL setups, Bayesian FL tools implemented over decentralized networks are subject to high communication costs due to the iterated exchange of local posterior distributions among cooperating devices.
Therefore, this paper proposes a communication-efficient decentralized Bayesian FL policy to reduce the communication overhead without sacrificing final learning accuracy and calibration. 
The proposed method integrates compression policies and allows devices to perform multiple optimization steps before sending the local posterior distributions. 
We integrate the developed tool in an Industrial Internet of Things (IIoT) use case where collaborating nodes equipped with autonomous radar sensors are tasked to reliably localize human operators in a workplace shared with robots.
Numerical results show that the developed approach obtains highly accurate yet well-calibrated ML models compatible with the ones provided by conventional (uncompressed) Bayesian FL tools while substantially decreasing the communication overhead (i.e., up to 99\%).
Furthermore, the proposed approach is advantageous when compared with state-of-the-art compressed frequentist FL setups in terms of calibration, especially when the statistical distribution of the testing dataset changes.   
\end{abstract}


\section{Introduction}
\label{sec:introduction}

Nowadays, \ac{fl} is used to obtain high-quality \ac{ml} models that are trained from decentralized data sources without disclosing any private information~\cite{advances_fl,fl_spm, fl_magazine, fl_camad}.
Given the privacy-preserving features of \ac{fl}, several industrial applications have integrated it to provide enhanced learning functionalities compared to conventional data center-based training strategies. 
Examples of such applications include \ac{iiot} services~\cite{fl_iiot,fl_iiot2}, autonomous driving use cases~\cite{fl_vehicle1,fl_access}, and healthcare~\cite{fl_healthcare,fl_bernardo}.
Still, most of the studied \ac{fl} implementations focus on standard (frequentist) strategies, where the goal is to find a single (optimized) set of \ac{ml} model parameters that best fit the training data. 
Following such a strategy has been shown to produce models that output overconfident predictions regardless of their correctness, especially when the local datasets of the devices are limited in size~\cite{simeone2022machine,channel_driven_bfl}.
\textcolor{black}{This raises major safety concerns for industrial services as downstream tasks may rely on the overconfident and often incorrect output provided by the \acp{nn}.}
To overcome this limitation, Bayesian \ac{fl} tools \cite{bayesian_fl_survey} target the learning of the posterior distribution in the model parameter space. 
By doing so, a reliable uncertainty measure can be obtained and used to make more informed decisions, consequently improving safety. 

Bayesian \ac{fl} tools are typically implemented considering approximate methods as obtaining the full posterior is often intractable \cite{simeone2022machine}.
%(e.g., due to the large number of trainable parameters of the \acp{nn}).
The first class of techniques relies on \ac{vi} methodologies~\cite{pvi,d-svgd} 
where a surrogate distribution is learned in place of the true posterior, while the second uses \ac{mcmc} approaches~\cite{d-sgld,d-sgmcmc} that approximate the posterior density via random samples. 
Focusing on \ac{mcmc} approaches, some distributed implementations have been developed for centralized \cite{d-sgmcmc,w-flmc} or fully distributed network topologies \cite{channel_driven_bfl,d-sgld}, while assuming uncompressed communications.
%while assuming devices sharing the local model distributions in an uncompressed manner. 
More advanced designs have been introduced recently to reduce the communication footprint in Bayesian \ac{fl} setups.
For example, in \cite{quantized_langevin}, a communication-efficient Bayesian \ac{fl} policy is developed where devices compress their local gradients and apply variance reduction techniques to improve learning performances. 
Similarly, authors in \cite{elf_langevin} propose several compression operators for Langevin-based \ac{fl} strategies that rely on primal, dual, and bidirectional compression.
Other approaches instead quantize the gradients exchanged during the learning process with 1-bit compression~\cite{bayesian_fl_networks} or propose new Lagevin schemes that support compressed communications~\cite{bayesian_fl_sampling}.
Despite this recent progress, all the aforementioned techniques focus on centralized strategies while completely overlooking decentralized Bayesian \ac{fl} tools.
%implemented over decentralized networks. 

\textbf{Contributions.} The paper proposes a novel communication-efficient Bayesian \ac{fl} strategy suitable for fully decentralized industrial setups, referred to as \ac{cd-bfl}.
In industrial contexts where safety is paramount, Bayesian \ac{fl} tools are needed to obtain reliable \ac{ml} models that support trustworthy predictions.
Compared to previous works that rely on centralized architectures, this is the first work that develops a fully decentralized method integrating compressed communications among devices. 
%To the best of our knowledge, this is the first work that considers compressed communication over decentralized networks.
The proposed tool draws inspiration from \cite{choco-sgd} and extends it to support Bayesian \ac{fl} strategies based on Langevin dynamics approaches. 
Specifically, \ac{cd-bfl} integrates compression policies and allows devices to perform multiple optimization steps before sending updates to their neighbors, massively reducing the communication cost. 
Numerical results focus on an \ac{iiot} learning problem where devices aim at collaboratively localizing human operators inside a human-robot shared workspace. 
They show that \ac{cd-bfl} substantially reduces the communication overhead (i.e., roughly by 99\%) compared to a conventional uncompressed decentralized Bayesian \ac{fl} strategy without impacting final learning performances or calibration.
The proposed strategy is also advantageous at reliably quantifying the uncertainty when the statistical distribution of the testing data changes, while a state-of-the-art (compressed) frequentist \ac{fl} policy fails at producing well-calibrated models.  
  
The paper is organized as follows. 
Sec. \ref{sec:system_model} introduces the system model and reviews Langevin-based sampling algorithms and their federated variants, while Sec. \ref{sec:proposed_fl_strategy} presents the developed decentralized Bayesian \ac{fl} policy. 
Sec. \ref{sec:case_study} introduces the industrial case study used for evaluating the performances of \ac{cd-bfl} (Sec. \ref{sec:numerical_results}).
Finally, Sec. \ref{sec:conclusions} draws some conclusions.
\section{System Model and Bayesian learning preliminaries}
\label{sec:system_model}

\begin{figure}
    \centering
    \includegraphics[width=\linewidth]{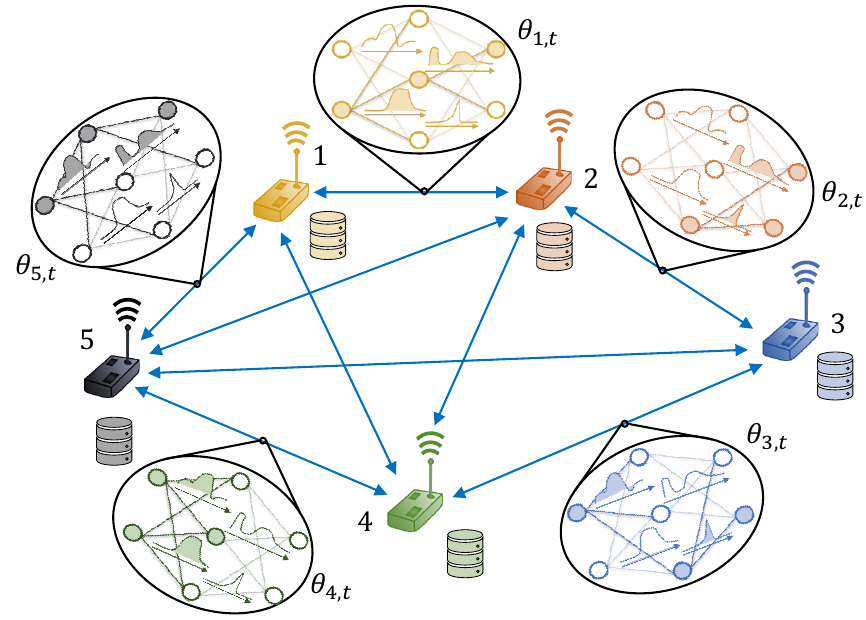}
    \vspace{-7mm}
    \caption{Bayesian \ac{fl} setup: collaborating nodes ($N = 5$) iteratively exchange compressed samples via \ac{d2d} communications and perform local computations to obtain a close approximation of the global posterior distribution. }
    \label{fig:bayesian_fl_setup}
\end{figure}

We focus on the decentralized \ac{fl} setup sketched in Fig.~\ref{fig:bayesian_fl_setup} where a set of devices $\mathcal{K} = \{1, \ldots, K\}$ collaborate for solving a supervised learning problem. %with $C$ possible output classes. 
Devices are connected according to the undirected graph $\mathcal{G} = (\mathcal{K}, \mathcal{E})$, with $\mathcal{E}$ being the set of directed edges. 
The set of neighbors of device $k$ including $k$  is denoted with $\mathcal{N}_k$, while the same set that excludes $k$ is indicated with $\mathcal{N}_{\overline{k}}$. %, while the same set that does not include $k$ is indicated with $\mathcal{N}_{\overline{k}}$. 
Each device $k \in \mathcal{K}$ is able to share information with its neighbors via \ac{d2d} communications and holds a local dataset $\mathcal{D}_k = \{(\mathbf{x}_{h}, y_h)\}_{h = 1}^{E_k}$ of $E_k$ examples pairs $(\mathbf{x}_{h}, y_h)$, with $\mathbf{x}_h$ and $y_h$ being the input data and desired prediction, respectively.
The global dataset is denoted with $\mathcal{D} = \{\mathcal{D}_k \}_{k = 1}^{N}$ and comprises $E = \sum_{k = 1}^{K} E_k$ training examples. 
As commonly observed in \ac{fl} setups, the local datasets may vary in size across different learners or under-represent the global dataset as only a limited number of classes is available at each device.
The devices' goal is to obtain a set of global model parameters $\boldsymbol{\theta} \in \mathbb{R}^{p}$ approximately sampled from the true global posterior $p(\boldsymbol{\theta} | \mathcal{D})$ of the model $\boldsymbol{\theta}$ which takes into account all the data available at the learners.
%The posterior $p(\boldsymbol{\theta} | \mathcal{D})$ should be obtained by the devices relying only on local computations and on \ac{d2d} communications for exchanging the model distributions. 
In what follows, the proposed Bayesian learning tools rely on gradient-based \ac{mcmc} approaches.
Specifically, we review \ac{sgld}, and then discuss its extension to fully decentralized networks.

\subsection{Centralized Stochastic Gradient Langevin Dynamics}

\ac{sgld}~\cite{sgld} aims at learning the posterior distribution in the model parameter space. 
To do so, each device shares its local dataset with a centralized unit in charge of carrying out the learning process.
Specifically, the goal is to obtain a close approximation of the global posterior distribution 
\begin{equation}\label{eq:post}
    p(\boldsymbol\theta | \mathcal{D}) \propto p(\boldsymbol\theta) \prod_{k = 1}^{K} p(\mathcal{D}_k | \boldsymbol\theta) \, , 
\end{equation}
where $p(\mathcal{D}_k | \boldsymbol\theta) = \prod_{h = 1}^{E_k} p(y_h | \mathbf{x}_h, \boldsymbol\theta)$ denotes the likelihood function describing the shared \ac{ml} model adopted by the learners, while $p(\boldsymbol{\theta})$ is the prior distribution.
Starting from the initial samples $\boldsymbol{\theta}_{0} \in \mathbb{R}^{p}$ at $t = 0$, \ac{sgld} produces new samples iteratively by adding Gaussian noise $\boldsymbol\xi_{t}$ to standard gradient descent updates as~\cite{sgld}
\begin{equation}
    \boldsymbol\theta_{t+1} = \boldsymbol\theta_t - \eta \nabla f(\boldsymbol\theta_{t}) + \sqrt{2\eta} \boldsymbol\xi_{t+1} \, , 
    \label{eq:sgld}
\end{equation}
where $t = 0, \ldots, T$ denotes the iteration number, $\boldsymbol\xi_{t+1}$ is a random vector drawn from an independent and identically (i.i.d.) Gaussian distribution $\mathcal{N}(\mathbf{0}_{p}, \mathbf{I}_{p})$, $\eta$ is the learning rate, while $f (\boldsymbol\theta_t) = \sum_{k = 1}^{N} f_k(\boldsymbol\theta_t, \mathcal{M}_k)$ with   
\begin{equation}
    f_k(\boldsymbol\theta_t, \mathcal{M}_k) = -\log p(\mathcal{M}_k | \boldsymbol{\theta}_t) - \dfrac{1}{N} \log p(\boldsymbol\theta_t) \, ,  \label{eq:gradient}
\end{equation} 
where $p(\mathcal{M}_k|\boldsymbol{\theta}_t)$ is the likelihood evaluated over a mini-batch $\mathcal{M}_k = \{(\mathbf{x}_b, y_b)\}_{m = 1}^{M}$ comprising $M$ training examples. 
%and $\boldsymbol\xi_{t+1}$ is a random vector drawn from an independent and identically (i.i.d.) Gaussian distribution $\mathcal{N}(\mathbf{0}_{m}, \mathbf{I}_{m})$, independent of the initial samples $\boldsymbol\theta_0$.
In practice, a burn-in phase is typically utilized in gradient-based \ac{mcmc} methods where the first $T_b$ samples are discarded while the remaining $T - T_b$ ones are used for uncertainty quantification.

\subsection{Decentralized Stochastic Gradient Langevin Dynamics}

\ac{sgld} is not directly applicable to \ac{fl} setups as all the data collected by the learners need to be sent to a centralized location, thereby raising privacy concerns. 
To overcome this limitation, \ac{dsgld}~\cite{d-sgld} has been introduced, which enables to implement \ac{sgld} over decentralized wireless networks without sharing any privacy-sensitive data. 
%More advanced designs have been therefore introduced where \ac{sgld} is implemented over wireless networks while keeping the data local. 
Under \ac{dsgld}, each device $k \in \mathcal{K}$ updates its local samples as~\cite{d-sgld}
\begin{equation}
    \boldsymbol\theta_{k. t+1} = \sum_{j \in \mathcal{N}_k} \omega_{kj} \boldsymbol\theta_{j,t} - \eta \nabla f_k(\boldsymbol\theta_{k, t}, \mathcal{M}_k) + \sqrt{2 \eta} \boldsymbol\xi_{k, t} \, , \label{eq:dsgld}
\end{equation} 
where $\omega_{k,j}$ is the $(k,j)$-th entry of a symmetric, doubly stochastic $K \times K$ matrix $\boldsymbol\Omega$ which can be chosen according to the choices presented in~\cite{fast_consensus}. 
%Following \eqref{eq:dsgld}, each agent firstly combines its local samples $\boldsymbol{\theta}_{j, t}$ with the ones received from its neighbors $j \in \mathcal{N}_k$ while also applying a noisy gradient descent update as in \eqref{eq:sgld}. 
%\ac{dsgld} has been proven to converge to the global posterior distribution defined in \eqref{eq:post} provided that the learning rate is properly tuned, the $\mathcal{G}$ is connected, and the local functions $f_k(.)$ are smooth and strongly convex. 
Still, \ac{dsgld} entails excessive communication overhead to exchange the samples among neighboring devices.
To address this shortcoming, we propose a communication-efficient \ac{dsgld} implementation, as detailed in the next section. 

% \begin{equation}
%     \nabla f_k (\boldsymbol\theta_{k}^{[s]}) = - \nabla \log p(\mathcal{D}_{k} | \boldsymbol\theta_k^{[s]}) - \dfrac{1}{N} \nabla \log p( \boldsymbol\theta_k^{[s]}) \, ,
% \end{equation}
% is the local gradient. 
%Similarly, \ac{d-sghmc} \cite{JMLR:v22:21-0307} lets each agent $k$ update its local samples as
%\begin{align}
%    \boldsymbol\nu_k^{[s+1]} &= \boldsymbol\nu_k^{[s]} - \eta [\gamma \boldsymbol\nu_k^{[s]} + \nabla f_k(\boldsymbol\theta_k^{[s]})] + \sqrt{2 \gamma \eta} \boldsymbol\xi_k^{[s+1]} \, ,  \\ 
%    \boldsymbol\theta_k^{[s+1]} &= \sum_{j \in \mathcal{N}_k} w_{kj} \boldsymbol\theta_j^{[s]} + \eta \boldsymbol\nu_k^{[s+1]} \, ,
%    \label{eq:dsgmhc}
%\end{align}
%\begin{equation}
%    \boldsymbol\theta_{k}^{[s+1]} = \sum_{j \in \mathcal{K}_{k}}w_{kj}\boldsymbol\theta_{j}^{[s]} + (1 - \gamma)\eta \boldsymbol\nu_{k}^{[s]}  - \eta \nabla f_k(\boldsymbol\theta^{[s]}) + \sqrt{2 \gamma \eta} \boldsymbol\xi_{k}^{[s+1]} \, , \label{eq:dsgmhc}
%\end{equation}

\section{Compressed Bayesian FL strategy}
\label{sec:proposed_fl_strategy}

\begin{algorithm}[!t]
\caption{CD-BFL}\label{alg:bayesian_strategy}
\textbf{Input:} initial samples $\boldsymbol{\theta}_{k,0}\, \forall \,  k \in \mathcal{N}$, graph $\mathcal{G}$, matrix $\boldsymbol{\Omega}$, mixing weight $\zeta$,  initialize $\mathbf{v}_{k,0} = \mathbf{0}_{p}$ and $\bar{\mathbf{v}}_{j,0} = \mathbf{0}_p$ $\forall \, j \in \mathcal{N}$
\begin{algorithmic}[1]
\For{each round $t = 0,1, \ldots T$}
\For{each learner $k \in \mathcal{N}$}
    \State $\boldsymbol{\theta}_{k,t}^{(0)} \leftarrow \boldsymbol{\theta}_{k,t}$
    \For{each $\ell = 1, \ldots, L$}
        \State $\boldsymbol{\theta}_{k,t}^{(\ell)} = \boldsymbol{\theta}_{k,t}^{(\ell-1)} - \eta \nabla f_k\left(\boldsymbol\theta_{k,t}^{(\ell-1)}, \mathcal{M}_{k}^{(\ell)}\right)$
    \EndFor 
    \State $\Delta \boldsymbol{\theta}_{k,t} = \mathcal{Q}\left(\boldsymbol{\theta}_{k,t}^{(L)}-\mathbf{v}_{k,t} \right)$ %\Comment{Compression}
    \State send $\left(\Delta \boldsymbol{\theta}_{k,t}\right)$ and receive $\{\Delta \boldsymbol{\theta}_{j,t}\}_{j\in\mathcal{N}_{\overline{k}}}$ 
    %\State receive $\{\mathbf{p}_{t,j}\}_{j\in\mathcal{N}_{\overline{i}}}$ 
    \State $\mathbf{v}_{k,t+1}=\mathbf{v}_{k,t}+ \Delta\boldsymbol{\theta}_{k,t}$
    \State $\bar{\mathbf{v}}_{k,t+1}=\bar{\mathbf{v}}_{k,t}+\sum_{j\in\mathcal{N}_{i}}\omega_{kj} \, \Delta \boldsymbol{\theta}_{j,t}$
    \State $\boldsymbol{\theta}_{k,t+1}=\boldsymbol{\theta}_{k,t}^{(L)}+\zeta(\bar{\mathbf{v}}_{k,t+1}-\mathbf{v}_{k,t+1}) + \sqrt{2\eta} \xi_{k,t+1}$ 
    
\EndFor
\EndFor
\end{algorithmic}
\end{algorithm}

This section presents the proposed communication-efficient decentralized Bayesian \ac{fl} strategy, namely \ac{cd-bfl}, which is summarized in Algorithm \ref{alg:bayesian_strategy}. 
\ac{cd-bfl} draws inspiration from~\cite{sgld} and introduces compression strategies to reduce the communication overhead required to implement \ac{dsgld} over wireless networks. Besides, it allows devices to perform multiple gradient descent steps before each communication phase, improving reliability as shown in Sec.~\ref{sec:numerical_results}. The proposed scenario is critical in industrial setups, such as the one discussed in Sec. \ref{sec:case_study}, where networked devices have limited resources.

Let us denote the current model iterate $\boldsymbol{\theta}_{k,t}$ available at device $k$ at iteration $t$, each device updates it recursively using stochastic gradient descent for $L$ local steps.
Specifically, starting from $\boldsymbol{\theta}_{k,t}^{(\ell)} = \boldsymbol{\theta}_{k,t}$ when $\ell = 0$, each step $\ell \geq 1$ with $\ell = 1, \ldots, L$ is evaluated as
\begin{equation}
    \boldsymbol{\theta}_{k,t}^{(\ell)} = \boldsymbol{\theta}_{k,t}^{(\ell -1)} - \eta \nabla f_k\left(\boldsymbol{\theta}_{k,t}^{(\ell-1)}, \mathcal{M}_k\right)
\end{equation}
where $\mathcal{M}_k$ is a mini-batch containing $M$  training examples (drawn randomly from the local dataset $\mathcal{D}_k$) while $\eta$ is the same learning rate defined in \eqref{eq:sgld}.
Then, instead of exchanging $\boldsymbol{\theta}_{k,t}$ as it is, the devices use an additional control sequence $\mathbf{v}_{k,t}$, with $\mathbf{v}_{k,0} = \mathbf{0}_{p}$ for $t = 0$, to preserve the average of the local iterates over consecutive iterations while also enforcing the noise introduced by the compression stage to vanish for a sufficiently high number of iterations $t$~\cite{choco-sgd}.
Given the current control sequence $\mathbf{v}_{k,t}$, each device computes the following (compressed) quantity 
\begin{equation}
    \Delta \boldsymbol{\theta}_{k,t} = \mathcal{Q}\left(\boldsymbol{\theta}_{k,t}^{(L)} - \mathbf{v}_{k,t} \right)
\end{equation}
where $\mathcal{Q}(.)$ denotes a compression operator (see e.g., \cite{qsgd} or \cite{top-k} for widely-used compression policies). 
The compressed representation $\Delta \boldsymbol{\theta}_{k,t}$ is then exchanged over the \ac{d2d} links. 

Devices receiving the compressed samples from their neighbors first update their control sequences as
\begin{equation}
    \mathbf{v}_{k,t+1} = \mathbf{v}_{k,t} + \Delta \boldsymbol{\theta}_{k,t}.    
\end{equation}
Next, they update an additional variable $\bar{\mathbf{v}}_{k,t}$, with $\bar{\mathbf{v}}_{k,0} = \mathbf{0}_p$ for $t = 0$, that stores the compressed representation of the control sequences received from their neighbors as
\begin{equation}
    \bar{\mathbf{v}}_{k,t+1} = \bar{\mathbf{v}}_{k,t} + \sum_{j \in \mathcal{N}_{k}} \omega_{kj} \Delta \boldsymbol{\theta}_{j,t}   
\end{equation}
where $\omega_{kj}$ is again the $(k,j)$-th entry of a symmetric, doubly stochastic matrix $\boldsymbol{\Omega}$. 
Finally, devices update their local model iterates via a consensus-based aggregation strategy while also adding the Gaussian noise as in standard \ac{dsgld}
\begin{equation}
    \boldsymbol{\theta}_{k,t+1} = \boldsymbol{\theta}_{k,t}^{(L)}+\zeta(\bar{\mathbf{v}}_{k,t+1}-\mathbf{v}_{k,t+1}) + \sqrt{2\eta} \boldsymbol\xi_{k,t+1}
\end{equation}
where $0 < \zeta \leq 1$ is a mixing parameter that can be optimized to improve learning performance or calibration. 
As in gradient-based \ac{mcmc} methods, the first $T_b$ samples $\{\boldsymbol{\theta}_{k,t} \}_{t = 0}^{T_b}$ are discarded while the remaining $T - T_b$ ones, namely $\{\boldsymbol{\theta}_{k,t} \}_{t = T_b+1}^{T}$, are used for evaluating the performances.

\section{Case study: HRC workspace}
\label{sec:case_study}

\begin{figure}
    \centering
    \includegraphics[width=0.99\linewidth]{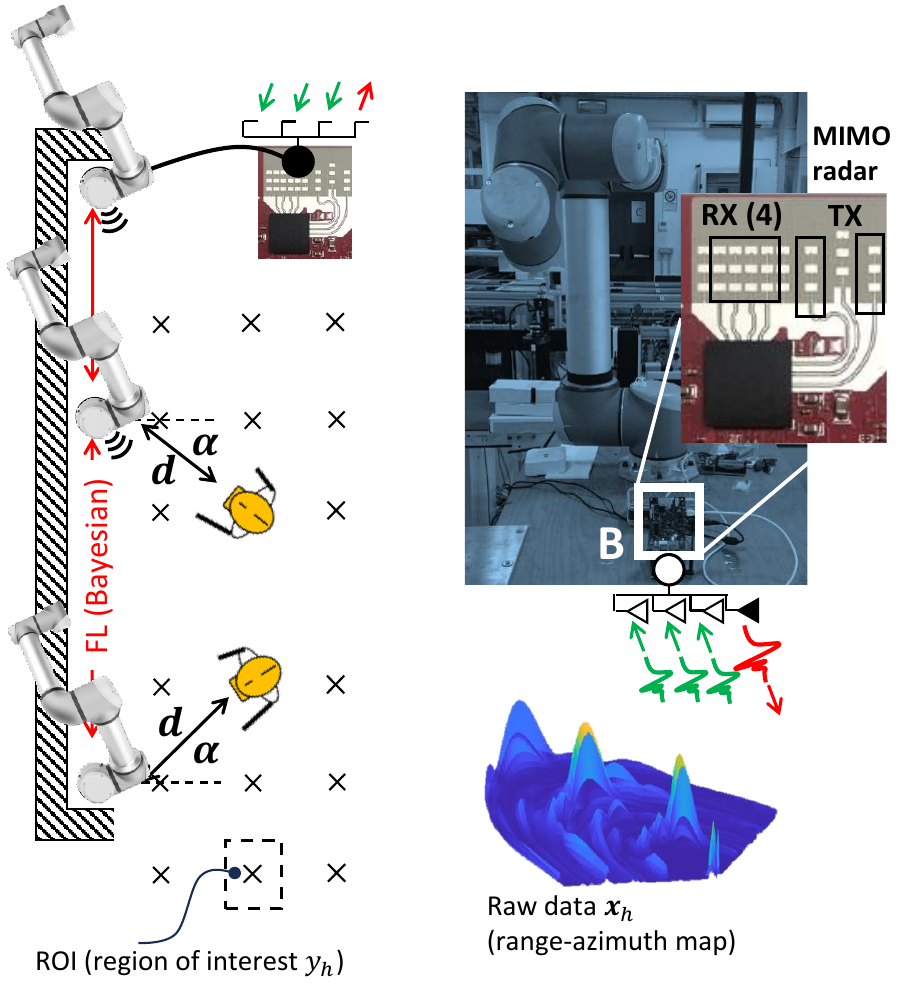}
    \vspace{-8mm}
    \caption{Human-Robot-Cooperative workspace scenario and FL setup.}
    \label{fig:hrc_setup}
\end{figure}

This section discusses the proposed Bayesian federation model \ac{cd-bfl} based
on experimental data. As depicted in Fig. \ref{fig:hrc_setup}, the proposed
FL scenario resorts to a network of radar devices \cite{cloud0} each
equipped with a Time-Division Multiple-Input-Multiple-Output (TD-MIMO)
Frequency Modulated Continuous Wave (FMCW) radar working in the 77\textminus 81
GHz band \cite{fmcw}.
The above-cited devices are employed to monitor
a shared industrial workspace during Human-Robot Collaboration (HRC)
tasks to detect and track the position of the human operators (the range distance $d$ and the Direction Of Arrival (DOA) $\alpha$)
relative to a robotic manipulator inside a fenceless space \cite{robot}.
In industrial shared workplaces, measuring positions and distance
is mandatory to enforce a worker protection policy and to implement collision
avoidance \cite{cobot}. 

Each radar features an array of $3$ TX and $4$ uniformly spaced
RX antennas, with an azimuth Field Of View (FOV) of -/+ $60$°, with angle
and range resolution of $25$° and $4.2$cm, and 3.9 GHz band (sweep time
$36$$\mu$s). Radars autonomously compute the beat signals on each
receiving antenna as the result of the radar echoes reflected by
moving objects. Beat signals are then converted in frequency domain
via Fast Fourier Transform (FFT) to obtain range-azimuth map measurements
$\mathbf{x}_{h}$. Maps $\mathbf{x}_{h}$ serve as training data and
have size $256$ x $63$ samples. Radars use a trained ML model to obtain the relative position ($d$, $\alpha$) information.
In addition, the subject position can be sent to a programmable logic
controller for robot safety control, for emergency stop or replanning
tasks. 

The ML model is here trained to detect the human subject in $R=10$
Region Of Interests (ROIs), namely potential collaborative situations
characterized by different human-robot (HR) distances or DOAs. In
particular, the ROI with label $y_{h}=0$ corresponds to the robot
and the corresponding human worker cooperating at a safe distance
(distance $\geq2$ m), the remaining $9$ labels ($y_{h}>0$) identify
the human operator as working close by the robot, at variable distances
and DOAs as depicted in Table \ref{labels}. The range-azimuth measurements
$\mathbf{x}_{h}$ are collected independently by the individual devices
using a dedicated radar DSP (C674x). The corresponding labels $y_{h}$
are associated manually during training stages and stored locally.
Federated model training is then implemented to replace data fusion
and using an ARM-Cortex-A57 SoC (Jetson Nano device model). Notice
that the radars collect a large amount of data, that cannot be shipped
back to the server for training and inference, due to the latency
constraints imposed by the localization service and safety policies
\cite{robot}.

\begin{table}[tp]
\medskip{}
 \protect\caption{\label{labels}Labels (HR collaborative situations) for $R=10$ ROIs.}
\vspace{-0.2cm}
\begin{centering}
\begin{tabular}{l|l|l|}
Labels & ROI: range $d$ & \multicolumn{1}{l}{ROI: DOA $\alpha$}\tabularnewline
\hline 
$y_{h}=0$ & $d\geq2$m & $-60\leq\alpha\leq60$ deg\tabularnewline
\hline 
$y_{h}=1$ & $0.5\mathrm{m}\leq d\leq0.7\mathrm{m}$ & $40\leq\alpha\leq60$ deg\tabularnewline
\hline 
$y_{h}=2$ & $0.3\mathrm{m}\leq d\leq0.5\mathrm{m}$ & $-10\leq\alpha\leq10$ deg\tabularnewline
\hline 
$y_{h}=3$ & $0.5\mathrm{m}\leq d\leq0.7\mathrm{m}$ & $-60\leq\alpha\leq-40$ deg\tabularnewline
\hline 
$y_{h}=4$ & $1\mathrm{m}\leq d\leq1.2\mathrm{m}$ & $20\leq\alpha\leq40$ deg\tabularnewline
\hline 
$y_{h}=5$ & $0.9\mathrm{m}\leq d\leq1.1\mathrm{m}$ & $-10\leq\alpha\leq10$ deg\tabularnewline
\hline 
$y_{h}=6$ & $1\mathrm{m}\leq d\leq1.2\mathrm{m}$ & $-40\leq\alpha\leq-20$ deg\tabularnewline
\hline 
$y_{h}=7$ & $1.2\mathrm{m}\leq d\leq1.6\mathrm{m}$ & $10\leq\alpha\leq20$ deg\tabularnewline
\hline 
$y_{h}=8$ & $1.1\mathrm{m}\leq d\leq1.5\mathrm{m}$ & $-5\leq\alpha\leq5$ deg\tabularnewline
\hline 
$y_{h}=9$ & $1.2\mathrm{m}\leq d\leq1.6\mathrm{m}$ & $-20\leq\alpha\leq-10$ deg\tabularnewline
\hline 
\end{tabular}
\par\end{centering}
%\medskip{}
% \protect\caption{\label{labels}Labels (HR collaborative situations) for $L=10$ ROIs.}
%\vspace{-0.4cm}
\end{table}

The \ac{cd-bfl} tool has been simulated on a virtual environment which
creates an arbitrary number of virtual radar devices configured to
process an assigned batch of range-azimuth data $\mathbf{x}_{h}$
and exchanging compressed samples as detailed in Sec. \ref{sec:proposed_fl_strategy}. The model adopted for localization is based on a Lenet architecture~\cite{mnist} with $p = 2.7$ million trainable parameters and is also trained continuously to track the variations of data dynamics
caused by changes in the workflow processes (typically, on a daily
basis). The initial training of the ML model is obtained at day $i = 1$ using a large dataset $\mathcal{D}^{(1)} =\{(\mathbf{x}_h, y_h)\}_{h =1}^{E^{(1)}}$ of raw range-azimuth
data manually labeled. Possible re-training stages occur daily, $i>1$
and are based on new data. Datasets for
initial model training and two subsequent re-training stages, $\mathcal{D}^{(i)}$ with $i = 2,3$, i.e., two consecutive days, are available online in \cite{ieeedataport}.

\section{Numerical Results}
\label{sec:numerical_results}

\textcolor{black}{In this section, we evaluate \ac{cd-bfl} over a network of $K = 10$ radars each having $50$ independently and identically distributed (iid) range-azimuth maps taken from the dataset $\mathcal{D}^{(1)}$ \cite{ieeedataport}.} 
%The radars employ a LeNet model \cite{mnist} with $p = 2.7$ million trainable parameters for mapping the input data $\mathbf{x}_h$ to the desired output labels $y_h$.
Connectivity among radars is simulated and assumes that each radar communicates with all $K-1$ neighbors. 

As performance metrics, we utilize the standard measure of validation accuracy and the \ac{ece}~\cite{ece}, which quantifies the \ac{ml} model's reliability. % obtained by the aforementioned strategies. 
Specifically, the \ac{ece} measures the disagreement between the accuracy and the confidence scores provided by the \acp{nn} and is computed as follows. 
Given the predictions of the \ac{ml} model over the validation dataset, they are divided into a set of $O$ non-overlapping bins $\{B_{o}\}_{o = 1}^{O}$ according to their associated confidence scores.
These scores are computed by taking the maximum probability assigned by the softmax operation at the last layer of the \ac{nn}.
Then, we compute the average accuracy $\text{acc}(B_{o})$ and average confidence $\text{conf}(B_{o})$ for each bin separately.
%over the examples having confidence falling inside the limits of the bin. 
Finally, the \ac{ece} is evaluated by taking into account all bins as~\cite{ece}
\begin{equation}
\text{ECE} = \sum_{o=1}^{O} \dfrac{|B_o|}{\sum_{o' = 1}^{O} |B_{o'}|} |\text{acc}(B_o) - \text{conf}(B_o)| \, , 
\end{equation}
where $|B_{o}|$ denotes the number of validation samples falling inside the $o$-th bin. 

The experiments compare the performances of \ac{cd-bfl} with the ones attained by an uncompressed Bayesian \ac{fl} method, namely \ac{dsgld}, as well as the state-of-the-art frequentist \ac{fl} approach developed in \cite{choco-sgd}, which we refer to as \ac{cf-fl}.
All methods use a learning rate of $\eta = 10^{-4}$ and are trained for $T = 800$ iterations.
For the Bayesian \ac{fl} setups, we consider a standard Gaussian as prior, i.e., $p(\boldsymbol{\theta}) = \mathcal{N}(\mathbf{0}_{p}, \mathbf{I}_{p})$ and a burn-in phase comprising the first $T_{b} = 700$ iterations.
Top-$k$ sparsification~\cite{top-k} is used as compression operator by \ac{cf-fl} and \ac{cd-bfl} and is configured such that only $1\%$ of the model parameters is exchanged by the devices, allowing to reduce the communication overhead by $99\%$.
Finally, all methods set the entries of $\boldsymbol{\Omega}$ as in \cite{fl_osvaldo} and we select $\zeta = 0.03$. 

\subsection{Results}

\begin{figure}
    \centering
\subfloat[\label{fig:acc_comparison}]{\includegraphics[width=0.98\linewidth]{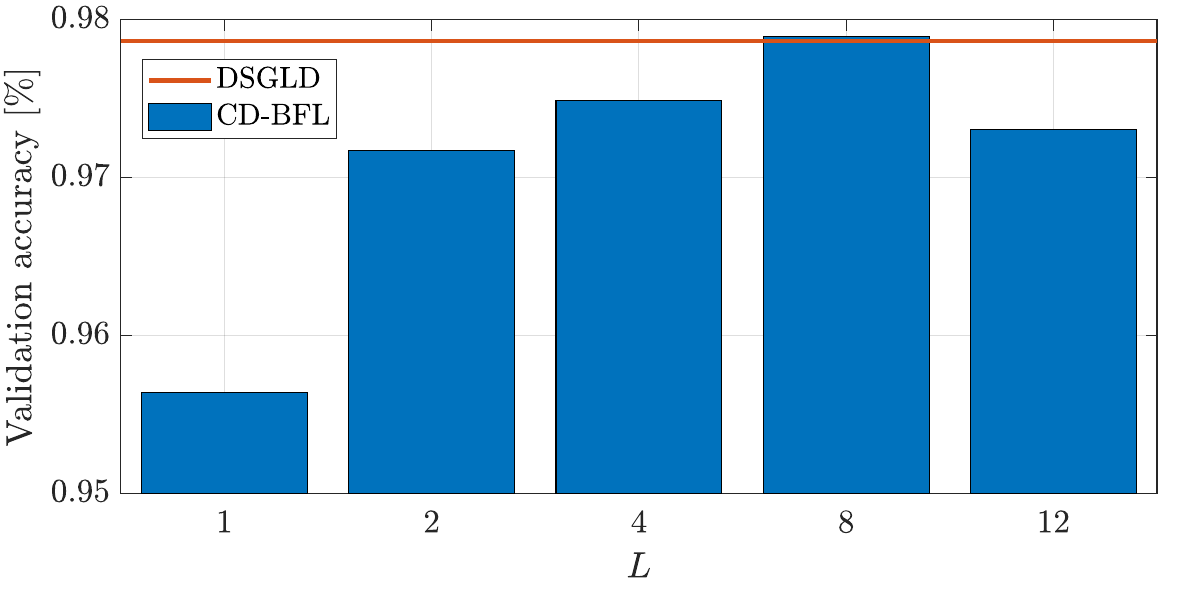}
	}

 \vspace{-3mm}
\subfloat[\label{fig:ece_comparison}]{\includegraphics[width=0.98\linewidth]{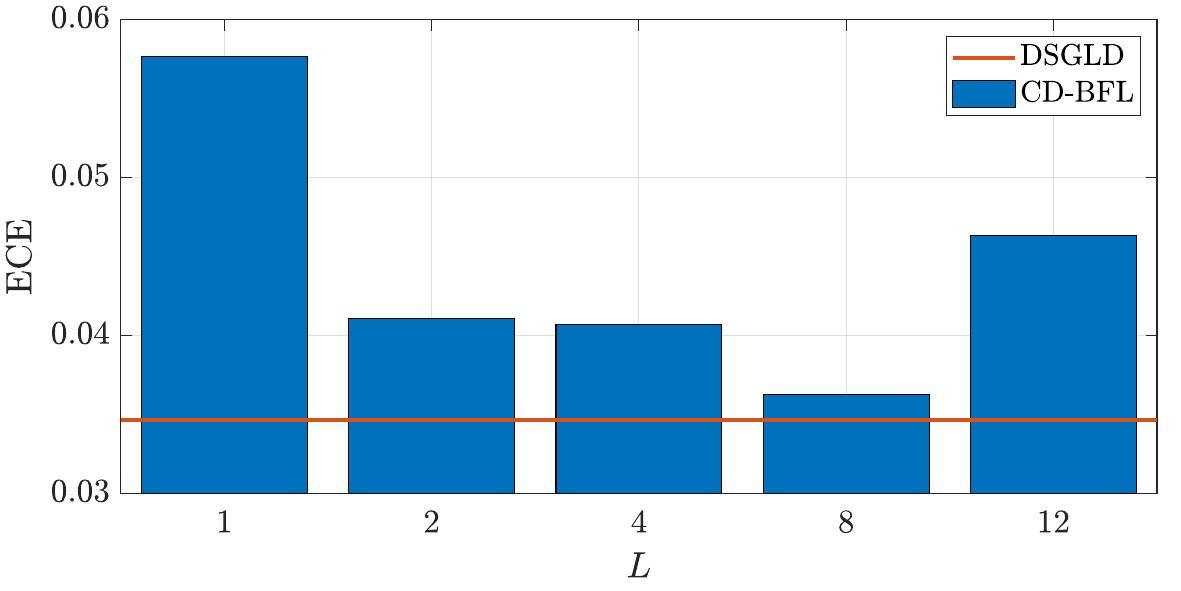}
	}
 \vspace{-4mm}
    \caption{Comparison between \ac{dsgld} and \ac{cd-bfl} for varying values of $L$ in terms of validation accuracy (a) and ECE (b). }
    \label{fig:comparision_bayesian}
\end{figure}

The first set of results evaluates the performances of \ac{cd-bfl} for varying number of local optimization steps $L$, ranging from $1$ up to $12$, and compares it with the ones attained by \ac{dsgld}.
Fig. \ref{fig:comparision_bayesian} reports the validation accuracy (Fig. \ref{fig:acc_comparison}) and the \ac{ece} (Fig. \ref{fig:ece_comparison}) achieved by \ac{cd-bfl} and \ac{dsgld}. 
Results show a trade-off between the number of local gradient descent updates $L$ and the final learning performances.
Indeed, increasing $L$ generally leads to better accuracy and \ac{ece} provided $L \leq 8$, while for $L > 8$ the models produced by the proposed strategy tend to be poorly calibrated and with limited generalization ability.
This should be expected as too many local optimization steps may introduce overfitting with a consequent reduction of the final accuracy and an increase in the \ac{ece}.
Therefore, in the considered setting, $L = 8$ should be chosen to maximize the prediction abilities and the reliability of the \acp{nn}.
By doing so, \ac{cd-bfl} attains nearly the same accuracy as \ac{dsgld} at the cost of a slightly higher \ac{ece}.

The second set of results aims at characterizing the ability of \ac{cd-bfl} to provide reliable models when the data collected by the radar devices have different statistical distributions (e.g., due to different radar configurations and/or slight changes in the HRC workspace).
To do so, we resort to the same dataset $\mathcal{D}^{(1)}$ in \cite{ieeedataport} where we train \ac{cd-bfl} with $L = 8$, as suggested by the previous analysis, \ac{dsgld}, and \ac{cf-fl} and we evaluate the learned models on the testing datasets of all other days.
In particular, the testing datasets are manipulated to contain only samples with labels $ 1\leq y_h \leq 6$ as considered highly critical due to the short distance (i.e., $d < 1.5$ m) between the robot and the human operator. 
Indeed, in those positions, reliable \ac{ml} models must be obtained so that robot control strategies can confidently use the \ac{nn} predictions to avoid injuries. 
%be utilized to avoid injuries caused by collisions between robots and human operators. 
In what follows, the results shown are averaged over the testing datasets of days $i = 2,3$.
%available in \cite{ieeedataport}.

\begin{figure}[!t]
    \centering
\subfloat[\label{fig:reliability_plot_bayesian}]{\includegraphics[width=0.98\linewidth]{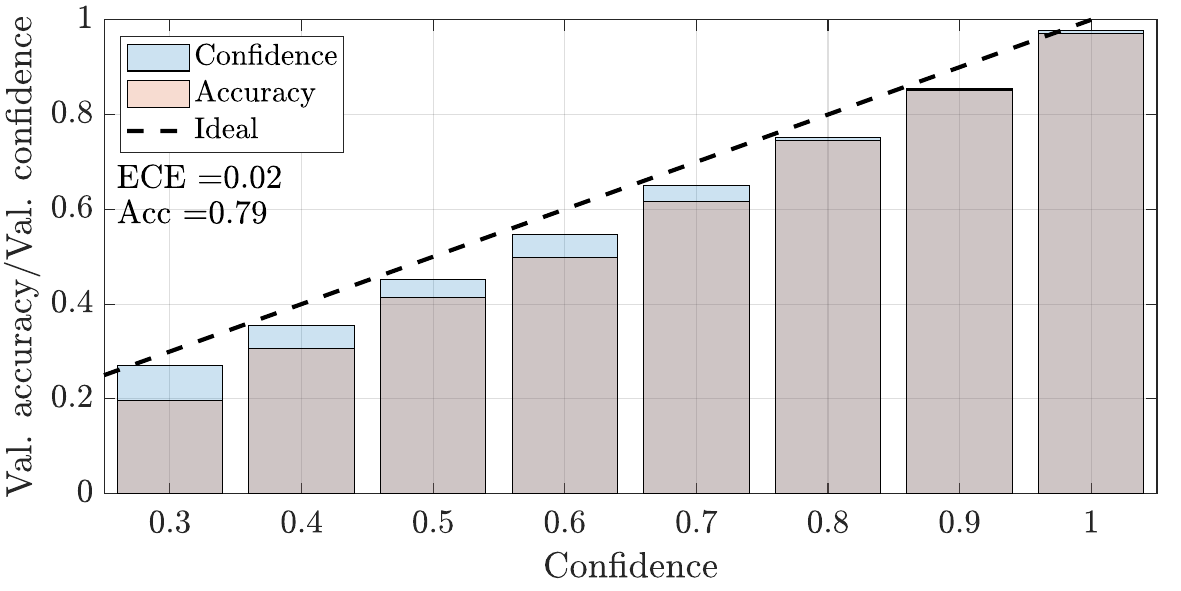}}

\vspace{-3mm}
\subfloat[\label{fig:reliability_plot_bayesian_compr}]{\includegraphics[width=0.98\linewidth]{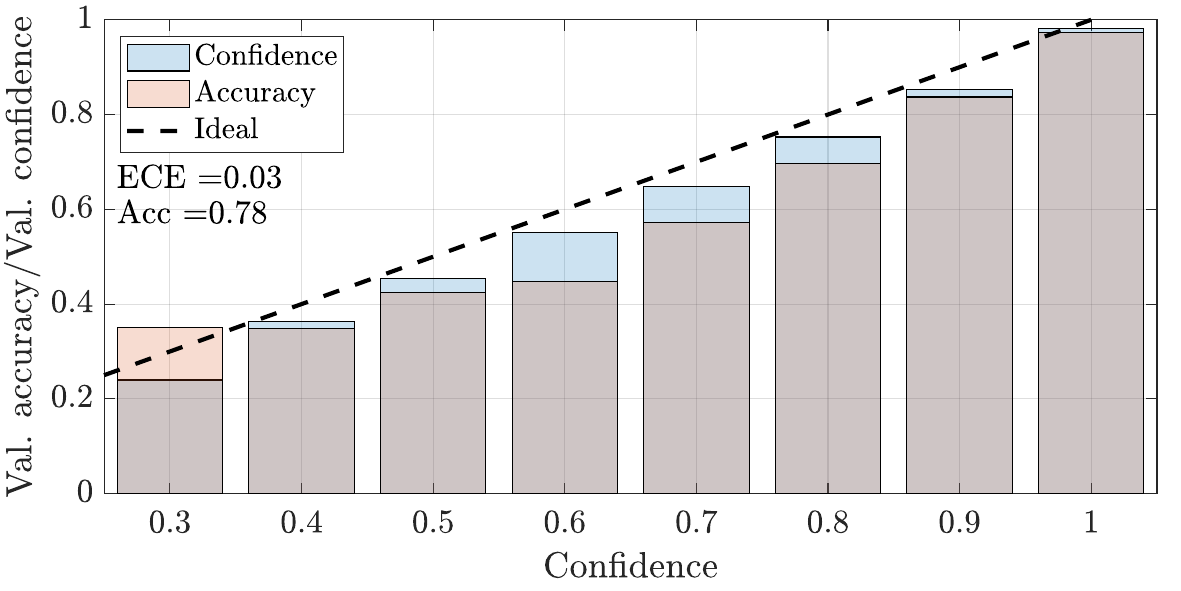}}

\vspace{-3mm}
\subfloat[\label{fig:reliability_plot_frequentist}]{\includegraphics[width=0.98\linewidth]{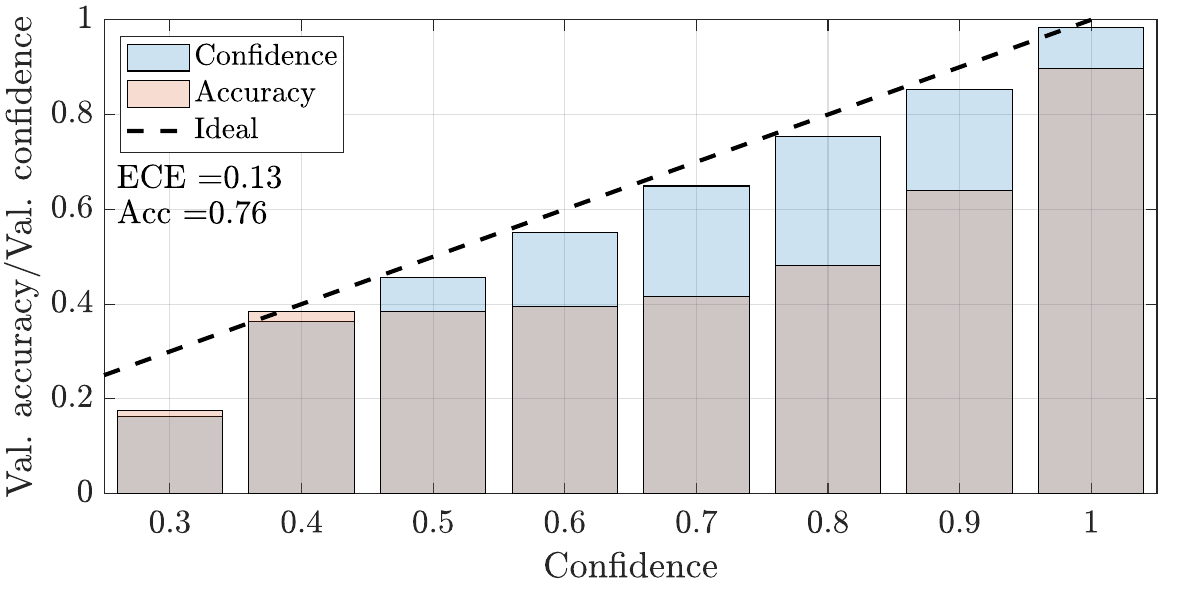}}
\vspace{-1mm}
    \caption{Reliability plots attained by (a) \ac{dsgld}, (b) \ac{cd-bfl} and (c) \ac{cf-fl}.}
    \label{fig:reliability_plots}
    \vspace{-3mm}
\end{figure}

Fig. \ref{fig:reliability_plots} reports the reliability diagrams~\cite{ece} which provide a visual representation of the reliability of the \ac{ml} models trained under \ac{dsgld} (Fig. \ref{fig:reliability_plot_bayesian}), \ac{cd-bfl} (Fig. \ref{fig:reliability_plot_bayesian_compr}) and \ac{cf-fl} (Fig. \ref{fig:reliability_plot_frequentist}) strategies. 
Comparing the results, even though \ac{cf-fl} and \ac{cd-bfl} provide the same savings in terms of communication overhead, they attain rather different reliability diagrams.  
Indeed, \ac{cf-fl} provides overconfident models since the confidence scores shown in Fig. \ref{fig:reliability_plot_frequentist} are much higher than the accuracy for most bins.
This poses major safety concerns as downstream tasks, e.g., robot control strategies, may rely on the (overconfident) predictions of the \ac{nn}, likely giving rise to injuries.
Besides, \ac{cf-fl} is shown to be the least accurate when compared with \ac{dsgld} or \ac{cd-bfl}. 
On the other hand, the reliability diagram of \ac{cd-bfl} shows that the confidence scores closely follow the actual accuracy. 
Therefore, the proposed strategy can be confidently used under safety-critical conditions as it provides well-calibrated \ac{ml} models able to reliably quantify the uncertainty of their predictions.
Still, some performance loss in terms of \ac{ece} and accuracy are exhibited by \ac{cd-bfl} compared to \ac{dsgld}, albeit this difference is very limited.

\section{Conclusions}
\label{sec:conclusions}

This paper proposed a communication-efficient decentralized Bayesian \ac{fl} policy, referred to as \ac{cd-bfl}, for reliable passive localization in \ac{iiot} setups. 
The developed method introduces compression operators while allowing devices participating in the learning process to perform multiple local optimization steps. The goal is to reduce communication overhead while maximizing the reliability of the \ac{ml} models. 
The proposed tool is integrated within a cooperative passive localization task where networked radars aim to accurately detect and classify human motions inside an industrial human-robot collaborative workplace. 
Numerical results show that \ac{cd-bfl} can be confidently utilized under safety-critical industrial operations as its \ac{ml} models reliably quantify the uncertainty of their predictions. 
Specifically, \ac{cd-bfl} provides accurate prediction capabilities as well as reliable uncertainty quantification that are in line with conventional uncompressed decentralized Bayesian \ac{fl} setups but at a much lower (i.e., 99\%) communication overhead.
Moreover, the models produced by \ac{cd-bfl} are well-calibrated even when they are used under different testing conditions, while state-of-the-art compressed decentralized \ac{fl} strategies fail at producing reliable \ac{ml} models. 

Future research works will target the theoretical characterization of the developed \ac{cd-bfl} tool and the development of strategies that jointly optimize the communication and the computational phases of the cooperative Bayesian \ac{fl} strategy.

\bibliographystyle{IEEEtran}
\bibliography{biblio.bib}

\end{document}